\documentclass[letterpaper, 10 pt, conference]{ieeeconf}

\IEEEoverridecommandlockouts
\usepackage{graphicx}
\usepackage{amsmath}
\usepackage{amssymb}
\usepackage{booktabs}
\usepackage{multirow}
\usepackage{tabularx}
\usepackage{xcolor}
\usepackage{caption}
\usepackage{placeins}
\usepackage{stfloats}
\usepackage{needspace}
\usepackage[hidelinks]{hyperref}

\newcommand{\figref}[1]{\hyperref[#1]{Fig.~\getrefnumber{#1}}}
\newcommand{\bestscore}[1]{\textcolor{red!75!black}{\textbf{#1}}}
\newcommand{\secondscore}[1]{\textcolor{blue!70!black}{\textbf{#1}}}
\newcommand{\methodblock}[1]{\par\noindent\textbf{#1.}\enspace\ignorespaces}
\newsavebox{\poseablationbox}
\makeatletter
\let\sure@normalsize\normalsize
\renewcommand{\normalsize}{\sure@normalsize
    \setlength{\baselineskip}{12pt plus 0.3pt minus 0.1pt}}
\renewcommand{\section}{\@startsection{section}{1}{\z@}
    {1.5ex plus 0.2ex minus 0.5ex}{0.7ex plus 0.1ex}
    {\normalfont\normalsize\centering\scshape}}
\renewcommand{\subsection}{\@startsection{subsection}{2}{\z@}
    {1.5ex plus 0.2ex minus 0.5ex}{0.7ex plus 0.1ex}
    {\normalfont\normalsize\itshape}}
\let\sure@orig@bibitem\@bibitem
\def\@bibitem#1{\sure@orig@bibitem{#1}\Hy@raisedlink{\hyper@anchorstart{cite.#1}\hyper@anchorend}}
\protected\def\cite#1{%
    \if@filesw\immediate\write\@auxout{\string\citation{#1}}\fi
    \begingroup
    \let\sure@cite@sep\@empty
    [\@for\sure@cite@key:=#1\do{%
        \sure@cite@sep
        \def\sure@cite@sep{,\penalty\@m\ }%
        \@ifundefined{b@\sure@cite@key}{%
            \textbf{?}\@warning{Citation `\sure@cite@key' on page \thepage\space undefined}%
        }{%
            \hyperlink{cite.\sure@cite@key}{\@nameuse{b@\sure@cite@key}}%
        }%
    }]%
    \endgroup
}
\makeatother

\title{SURE-Map: Self-Correcting Streaming Geometric Foundation Models}

\author{Mingkai Liu$^{1,2}$, Hao Zhao$^{3,*}$,
Xingxing Zuo$^{1,*}$
\thanks{$^{1}$ Mohamed bin Zayed University of Artificial Intelligence (MBZUAI), UAE.}%
\thanks{$^{2}$ Peking University, China.}%
\thanks{$^{3}$ Tsinghua University, China.}%
\thanks{$^*$Hao Zhao and Xingxing Zuo are the corresponding authors (Email: {\tt\small xingxing.zuo@mbzuai.ac.ae}).}
}

\begin{document}

\nocite{whelan2015elasticfusion,huang2023visual,hughes2022hydra,wang2025vggt,lin2025depth,wang2026pi,wang2024dust3r,leroy2024grounding,murai2025mast3r,maggio2026vggtslam2,deng2025vggt,zhang2026loger,xie2026scal3r,shen2025fastvggt,chen2026ttt3r,cheng2026longstream,wang2025continuous,lan2026stream3r,chen2026geometric,cheng2026horizonstream,tao2026oxford,geiger2012we,azinovic2022neural,schonberger2016structure,mur2015orb,von2018direct,forster2014svo,furukawa2009accurate,schonberger2016pixelwise,tyszkiewicz2020disk,lindenberger2023lightglue,duisterhof2025mast3r,teed2021droid,teed2023deep,zhuo2026streaming,yuan2026infinitevggt,li2026wint3r,rahaman2019spectral,wang2020tartanair,shotton2013scene,brizi2024vbr,oquab2023dinov2}

\maketitle
\thispagestyle{empty}
\pagestyle{empty}

\begin{abstract}
\textbf{
Streaming geometric foundation models are emerging as a compelling alternative to SLAM systems. Yet this streaming nature introduces a fundamental issue: each prediction is made from limited context, which is vulnerable to dynamic objects and weak textures. Small local errors accumulate into severe geometric distortion and long-horizon scale drift. We argue that reliable streaming reconstruction requires geometric foundation models to be not only predictive, but also self-correcting. We introduce SURE-Map, a self-correcting framework built upon two complementary principles. First, we explicitly model cross-view geometric uncertainty. Unlike conventional depth or point confidence, which primarily reflects the reliability of individual-view prediction, our uncertainty directly measures whether the jointly predicted pose and depth induce geometrically consistent cross-view pixel correspondences. Second, because local correction alone cannot eliminate slowly accumulating scale errors, we introduce multi-timescale self-correction: fast consecutive-frame inference preserves streaming efficiency, while sparse keyframe-window inference provides longer-range geometric evidence to periodically recalibrate the scale of recent trajectories. SURE-Map establishes new state-of-the-art performance for online feed-forward reconstruction across long-horizon benchmarks, reducing ATE-RMSE from 24.00 to 17.24 m on KITTI, 5.11 to 4.74 m on Oxford Spires, and 31.37 to 28.58 m on VBR, with further improvements to 15.17, 4.63, and 22.12 m when incorporating loop-closure refinement. Project page: \url{https://mingkai-liu.github.io/projects/sure-map/}.}

\end{abstract}

\section{Introduction}

3D reconstruction from video streams is a fundamental capability for augmented reality (AR)~\cite{whelan2015elasticfusion} and embodied intelligence~\cite{huang2023visual,hughes2022hydra}. Recent feed-forward geometric foundation models~\cite{wang2025vggt,lin2025depth,wang2026pi,wang2024dust3r,leroy2024grounding} have demonstrated a promising alternative to conventional SLAM pipelines by directly predicting camera poses and dense scene geometry from images, largely bypassing explicit feature matching, triangulation, and costly back-end optimization. However, extending these models to streaming reconstruction introduces a fundamental challenge. Unlike offline reconstruction, where predictions can leverage broad temporal context, an online system must operate causally with bounded memory and low latency, forcing each prediction to rely on only limited recent observations. Such restricted context makes the reconstruction particularly vulnerable to ambiguous local evidence, such as dynamic objects and weakly textured regions, where small errors in pose or geometry can progressively accumulate into severe geometric distortion and scale drift.

Existing approaches to long-horizon reconstruction largely trade off global geometric consistency against streaming efficiency. SLAM-hybrid systems~\cite{murai2025mast3r,maggio2026vggtslam2} integrate feed-forward geometric priors into classical mapping pipelines, using keyframe management, bundle adjustment, and pose-graph optimization to repeatedly enforce global consistency. While effective, such iterative back-end optimization introduces substantial computational overhead and weakens the simplicity and low latency of feed-forward inference. At the other extreme, offline feed-forward methods~\cite{deng2025vggt,zhang2026loger,xie2026scal3r,shen2025fastvggt} obtain longer-range geometric context by jointly processing complete sequences or reconstructing overlapping temporal chunks followed by global alignment. Their strong performance, however, relies on non-causal computation, large temporal context, or access to future frames, making them unsuitable for strict online reconstruction.

Recent state-of-the-art streaming reconstruction models LingBot-Map~\cite{chen2026geometric} and HorizonStream~\cite{cheng2026horizonstream} maintain compact geometric context for efficient online inference. Despite their practical memory-efficient design, these systems still inherit limitations from incremental pose estimation with limited context and remain vulnerable to observation ambiguities induced by dynamic objects, geometric degeneracy, weak textures, and repetitive structures. These ambiguities introduce local geometric inconsistencies that degrade dense point-cloud quality. Over long trajectories, the resulting local pose errors gradually accumulate into severe scale drift.

\begin{figure*}[t]
    \centering
    \makebox[\textwidth][c]{\includegraphics[width=1.03\textwidth,trim=4bp 4bp 5.5bp 2bp,clip]{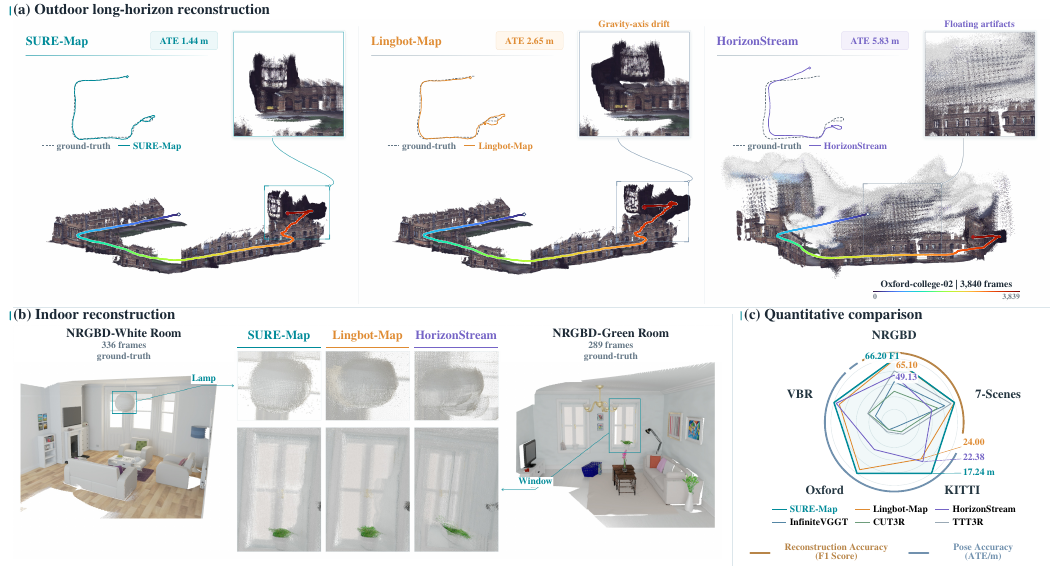}}
    \vspace{-0.6em}
    \caption{
        Comparison of SURE-Map with online feed-forward reconstruction methods~\cite{chen2026ttt3r,wang2025continuous,chen2026geometric,cheng2026horizonstream,yuan2026infinitevggt} across multiple benchmarks~\cite{azinovic2022neural,shotton2013scene,tao2026oxford,geiger2012we,brizi2024vbr}.
        (a) On long outdoor sequences, SURE-Map produces a more complete and accurate map; LingBot-Map exhibits severe drift along the gravity direction, while HorizonStream suffers from occlusion by floating artifacts.
        (b) SURE-Map reconstructs indoor windows and lamps with cleaner surfaces and sharper geometry, whereas LingBot-Map and HorizonStream exhibit visible artifacts.
        (c) Quantitative comparison of reconstruction accuracy (F1) and pose accuracy (ATE) across five benchmarks.
    }
    \label{fig:visual-summary}
    \vspace{-0.8em}
\end{figure*}

In this work, we propose SURE-Map (Scale- and Uncertainty-aware REconstruction), a self-correcting framework that equips streaming geometric foundation models with the ability to detect and rectify their own geometric failures. Our first key idea is to model cross-view geometric uncertainty. Instead of asking whether an individual depth or 3D prediction is reliable in isolation, SURE-Map estimates whether the jointly predicted pose and depth induce geometrically consistent pixel correspondences across consecutive views. This provides a direct measure of the reliability of the implicit data association underlying streaming reconstruction. However, such frame-to-frame correction alone cannot prevent small scale errors from accumulating over long trajectories. We therefore introduce multi-timescale self-correction, coupling fast consecutive-frame inference with sparse keyframe-window inference over a longer temporal horizon. The former preserves the low latency of streaming reconstruction, while the latter provides more stable long-range geometric evidence to periodically recalibrate the scale of recently estimated trajectories. \figref{fig:visual-summary} illustrates these complementary effects on indoor and outdoor scenes.

Our contributions are summarized as follows:
\begin{itemize}
\item We introduce \textbf{SURE-Map}, a self-correcting framework for streaming geometric foundation models, addressing the challenge that local geometric errors caused by limited temporal context can accumulate into severe reconstruction distortion and long-horizon scale drift.

\item We introduce \textbf{cross-view geometric uncertainty}, which reflects whether jointly predicted pose and depth from streaming models induce geometrically consistent pixel correspondences across views. Unlike conventional depth confidence that evaluates individual-view prediction, our formulation explicitly captures the reliability of cross-view geometry and enables uncertainty-aware geometric optimization.

\item We propose \textbf{multi-timescale self-correction}, combining efficient consecutive-frame inference (geometry-context-attention) with sparse keyframe-window inference (full-attention) to provide longer-range geometric evidence for periodic scale recalibration. This suppresses accumulated scale drift while preserving streaming efficiency, establishing new state-of-the-art performance on long-horizon benchmarks.

\end{itemize}

\section{Related Work}

\methodblock{Traditional 3D Reconstruction and Learned Back-ends} Classical 3D reconstruction pipelines include Structure-from-Motion (SfM)~\cite{schonberger2016structure}, Simultaneous Localization and Mapping (SLAM)~\cite{mur2015orb,von2018direct,forster2014svo}, and Multi-View Stereo (MVS)~\cite{furukawa2009accurate,schonberger2016pixelwise}. SfM and SLAM estimate camera motion and sparse structure through feature matching, keyframes, bundle adjustment, and pose-graph optimization, while MVS recovers dense geometry from posed images. These systems are accurate and interpretable, but their explicit matching and intensive optimization are costly for long-horizon streams. Recent methods incorporate learning through features and matchers~\cite{tyszkiewicz2020disk,lindenberger2023lightglue}, learned SfM/VO/SLAM modules~\cite{duisterhof2025mast3r,teed2021droid,teed2023deep}, and learned-prior back-ends~\cite{murai2025mast3r,maggio2026vggtslam2}, highlighting the growing role of learned 3D priors in reconstruction pipelines.

\methodblock{3D Foundation Models} Feed-forward 3D foundation models~\cite{wang2025vggt,lin2025depth,wang2026pi,wang2024dust3r,leroy2024grounding} have recently emerged as a strong alternative to optimization-heavy reconstruction pipelines. DUSt3R~\cite{wang2024dust3r} directly regresses dense point maps from unposed image pairs, but requires additional alignment to handle multiple views. VGGT~\cite{wang2025vggt} extends this paradigm to multi-view inputs, jointly predicting camera poses and dense 3D attributes in a single forward pass. Despite their strong accuracy and generalization, these models are designed for offline processing with full input access, and their full attention becomes costly on long-horizon streams.

\methodblock{Feed-Forward Streaming 3D Reconstruction} Feed-forward streaming 3D reconstruction extends geometric foundation models to causal video inputs. Recent methods~\cite{chen2026ttt3r,wang2025continuous,lan2026stream3r,zhuo2026streaming,yuan2026infinitevggt,li2026wint3r} transfer information across time through recurrent states, causal attention, sliding windows, keyframe memory, cache pruning, or test-time updates. Long-stream systems further improve context retention: LongStream~\cite{cheng2026longstream} analyzes degradation from attention sink and state saturation, HorizonStream~\cite{cheng2026horizonstream} mitigates scale drift via channel-wise geometric-evidence propagation, and LingBot-Map~\cite{chen2026geometric} improves long-range consistency with compact geometric context and large-scale long-sequence training. Despite improved scalability, bounded context leaves incremental pose estimation vulnerable to ambiguities that degrade dense reconstruction and induce local pose errors, which accumulate into long-trajectory scale drift. SURE-Map addresses this error propagation by using cross-view geometric uncertainty to identify unreliable observations and multi-timescale self-correction to correct trajectory errors across temporal scales.

\methodblock{Confidence in 3D Reconstruction} Per-pixel confidence is widely adopted in point-map prediction and depth estimation~\cite{wang2025vggt,lin2025depth,wang2026pi,wang2024dust3r}, while per-correspondence confidence is used in SLAM~\cite{murai2025mast3r,teed2021droid,teed2023deep}. Feed-forward 3D foundation models typically learn depth or point confidence via negative log-likelihood (NLL) regression losses and use it to filter unreliable geometry. These scores primarily reflect the reliability of individual-view geometry prediction, and their raw magnitudes need not be comparable across scenes. SURE-Map instead models cross-view geometric uncertainty by assessing whether jointly predicted pose and depth induce geometrically consistent pixel correspondences across consecutive views.

\section{Methodology}
\label{sec:method}

\subsection{Preliminary and Problem Definition}
\label{sec:problem-definition}

\methodblock{Causal Streaming Reconstruction} Given an image stream \(\{\mathbf{I}_0,\mathbf{I}_1,\ldots\}\), a streaming geometric foundation model processes each new frame upon arrival.
At time instant \(t_i\) capturing image \(\mathbf{I}_i\in\mathbb{R}^{H\times W\times 3}\), the model estimates the current camera pose and dense depth using the image and a compact memory summarizing historical observations:
\begin{equation}
    f_\theta(\mathbf{I}_i,\mathcal{M}_{i-1})
    \rightarrow
    \left(\mathbf{T}_i,\mathbf{D}_i,\mathbf{K}_i,\mathcal{M}_i\right),
    \label{eq:streaming-goal}
\end{equation}
where \(f_\theta\) denotes the streaming geometric foundation model, \(\mathbf{T}_i=[\mathbf{R}_i|\mathbf{t}_i]\in SE(3)\) is the camera-to-world pose, \(\mathbf{R}_i\in SO(3)\), \(\mathbf{t}_i\in\mathbb{R}^3\), \(\mathbf{D}_i\in\mathbb{R}^{H\times W}\) is the dense depth map, \(\mathbf{K}_i\) is the camera intrinsic matrix, and \(\mathcal{M}_{i-1}\) denotes the compact token memory summarizing historical observations until image \(\mathbf{I}_{i-1}\).

\methodblock{From Ambiguous Observations to Local Errors} Recent streaming geometric foundation models~\cite{cheng2026longstream,lan2026stream3r,chen2026geometric,cheng2026horizonstream} directly predict pose and dense depth without explicit 2D--2D correspondences, relying instead on implicit data association encoded by learned scene priors.
Neural networks favor smooth embeddings~\cite{rahaman2019spectral}, so visually similar but geometrically distinct observations may receive similar representations.
Non-causal offline systems~\cite{wang2025vggt,deng2025vggt,zhang2026loger,xie2026scal3r,shen2025fastvggt} can use future multi-view evidence to reject such spurious associations, whereas causal streaming inference is restricted to limited historical context.
It is therefore more vulnerable to ambiguities caused by dynamic objects, geometric degeneracy, weak textures, and repetitive structures, which degrade dense reconstruction and induce local pose errors.

\methodblock{From Local Errors to Scale Drift}
Unlike classical monocular SLAM, streaming geometric foundation models~\cite{cheng2026longstream,lan2026stream3r,chen2026geometric,cheng2026horizonstream} benefit from learned scene priors and often preserve an approximately stable scale within short segments.
However, causal streaming inference lacks explicit global cross-segment constraints to align the scales of different segments.
Consequently, local pose errors accumulate into segment-wise scale drift over long trajectories, even when individual segments remain locally plausible.
In contrast, dense depth estimation is largely driven by image appearance and tends to maintain a more stable local scale under the same limited context. SURE-Map leverages this favorable property through a dedicated inference procedure over a selected sparse keyframe window, periodically recalibrating the trajectory scale.

\subsection{Overview of SURE-Map}
\label{sec:method-overview}

\begin{figure*}[t]
    \centering
    \includegraphics[width=\textwidth]{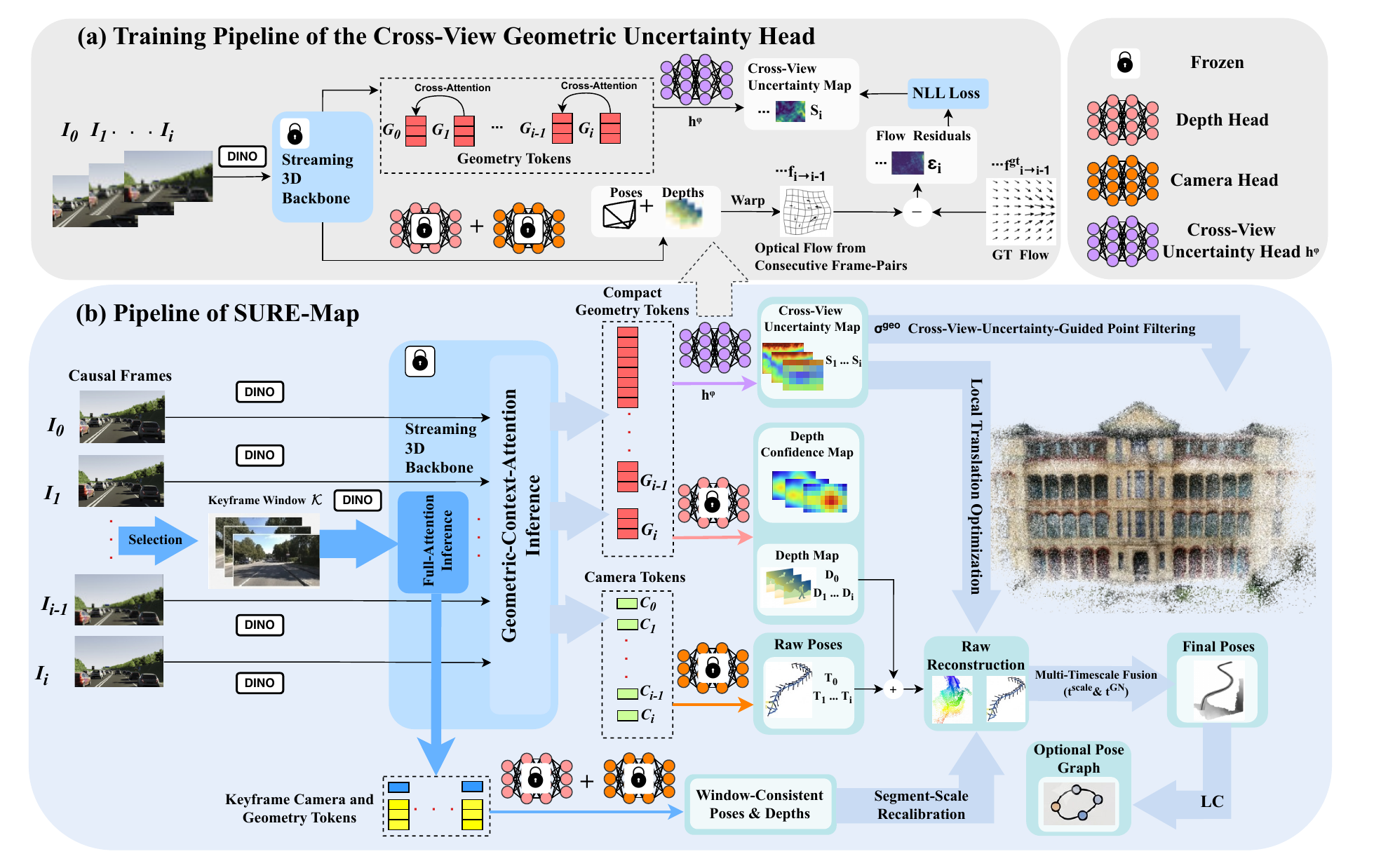}
    \vspace{-0.6em}
    \caption{
        SURE-Map methodology. (a) The cross-view geometric uncertainty head is supervised by discrepancies between pose-depth-induced and ground-truth optical flow. (b) During streaming reconstruction, the uncertainty guides dense-point filtering and local translation optimization. Multi-timescale self-correction couples fast consecutive-frame inference with sparse keyframe-window inference, whose longer-range geometric evidence periodically recalibrates trajectory scale.
    }
    \label{fig:pipeline}
    \vspace{-0.8em}
\end{figure*}

SURE-Map is a self-correcting framework for streaming geometric foundation models that addresses visual artifacts and segment-wise scale drift (\figref{fig:pipeline}).
First, cross-view geometric uncertainty assesses whether jointly predicted pose and depth induce geometrically consistent pixel correspondences across consecutive views, guiding dense-point filtering and local translation optimization.
Second, multi-timescale self-correction combines fast consecutive-frame inference (geometry-context-attention~\cite{chen2026geometric}) with sparse keyframe-window inference (full-attention), providing longer-range geometric evidence to periodically recalibrate the scale of recent trajectories.
Together, they improve dense geometry and long-horizon trajectory consistency while preserving streaming efficiency.

\subsection{Cross-View Geometric Uncertainty Modeling}
\label{sec:uncertainty-learning}

\methodblock{Pose-Depth-Induced Optical Flow} As shown in \figref{fig:pipeline}(a), given the pose and depth predicted by a streaming model, we can analytically derive the optical flow between two images. To quantify the cross-view geometric uncertainty in the predicted pose and depth, we augment the streaming model with a cross-view geometric uncertainty head. This head estimates uncertainty from geometric tokens and is trained under the supervision of the discrepancy between the ground-truth optical flow and the flow induced by the predicted pose and depth. During streaming inference, it directly estimates the uncertainty of cross-view correspondences \textit{without} explicitly predicting an optical-flow field.

For consecutive frames \((\mathbf{I}_{i-1}, \mathbf{I}_i)\) with resolution \(H \times W\), we predict a cross-view geometric uncertainty map \(\mathbf{S}_i\).
For any pixel \(\mathbf{u}=[u,v]^\top\) in the current frame \(\mathbf{I}_{i}\), with homogeneous coordinate \(\bar{\mathbf{u}}=[u,v,1]^\top\), we warp it into the previous frame \(\mathbf{I}_{i-1}\) using the predicted depth and relative pose.
The resulting backward optical flow \(\mathbf{f}_{i\rightarrow i-1}(\mathbf{u})\) represents the pose-depth-induced correspondence from \(\mathbf{I}_i\) to \(\mathbf{I}_{i-1}\):
\begin{equation}
\begin{array}{@{}l@{\;}c@{\;}l@{}}
    \mathbf{p}^{\mathrm{cur}}
    & = & \mathbf{D}_i(\mathbf{u})\mathbf{K}_i^{-1}\bar{\mathbf{u}},\\
    \mathbf{p}^{\mathrm{prev}}
    & = & \mathbf{R}_{i-1}^{\top}
    \left(\mathbf{R}_i\mathbf{p}^{\mathrm{cur}}
    +\mathbf{t}_i-\mathbf{t}_{i-1}\right),\\
    \mathbf{f}_{i\rightarrow i-1}(\mathbf{u})
    & = & \pi\!\left(\mathbf{K}_{i-1}\mathbf{p}^{\mathrm{prev}}\right)-\mathbf{u}.
\end{array}
    \label{eq:induced-flow}
\end{equation}
where \(\mathbf{p}^{\mathrm{cur}}\) is the 3D point back-projected from pixel \(\mathbf{u}\) in the current camera frame, \(\mathbf{p}^{\mathrm{prev}}\) is the same point expressed in the previous camera frame, and \(\pi(\cdot)\) denotes perspective division that projects 3D points onto the image plane.

\methodblock{Cross-View Geometric Uncertainty Prediction}
The head \(h_\psi\) takes the geometric token maps of two consecutive frames, \((\mathbf{G}_{i-1}, \mathbf{G}_i)\), as input.
These tokens encode local geometry and cross-view correspondence cues, enabling the head to predict the uncertainty associated with the pose-and-depth-induced backward optical flow. Specifically, the head outputs \(\mathbf{S}_i = h_\psi(\mathbf{G}_{i-1},\mathbf{G}_i) = [\mathbf{S}_i^u; \mathbf{S}_i^v] \in \mathbb{R}^{H \times W \times 2}\), where \(\mathbf{S}_i^u, \mathbf{S}_i^v \in \mathbb{R}^{H \times W}\) denote the horizontal and vertical log-variance maps, respectively.
For simplicity, \(\mathbf{S}_i(\mathbf{u})\) parameterizes only a diagonal covariance matrix:
\begin{equation}
\begin{array}{@{}l@{\;}c@{\;}l@{}}
    \boldsymbol{\Sigma}_i(\mathbf{u})
    =
    \operatorname{diag}\!\left(
        e^{\mathbf{S}_i^u(\mathbf{u})},
        e^{\mathbf{S}_i^v(\mathbf{u})}
    \right).
\end{array}
    \label{eq:geometric-uncertainty-covariance}
\end{equation}

\methodblock{Residual-Based Supervision} Given the ground-truth flow \(\mathbf{f}^{\mathrm{gt}}_{i\rightarrow i-1}\), the flow residual can be computed by
\(\boldsymbol{\epsilon}_i(\mathbf{u})=\mathbf{f}_{i\rightarrow i-1}(\mathbf{u})-\mathbf{f}^{\mathrm{gt}}_{i\rightarrow i-1}(\mathbf{u})\).
We train \(h_\psi\) over the set of pixels with valid optical flow, denoted by \(\Omega_i\), using the negative log-likelihood (NLL) loss:
\begin{equation}
\begin{aligned}
    \mathcal{L}_{\mathrm{flow}}
    &= \frac{1}{|\Omega_i|}\sum_{\mathbf{u}\in\Omega_i}\ell_i(\mathbf{u}),\\
    \ell_i(\mathbf{u})
    &=
    \boldsymbol{\epsilon}_i(\mathbf{u})^\top
    \boldsymbol{\Sigma}_i(\mathbf{u})^{-1}
    \boldsymbol{\epsilon}_i(\mathbf{u})
    +
    \log\!\left(\det\boldsymbol{\Sigma}_i(\mathbf{u})\right),
\end{aligned}
    \label{eq:flow-uncertainty-loss}
\end{equation}
In the NLL loss, the Mahalanobis term encourages the network to assign higher uncertainty to large residuals in the induced optical flow, whereas the log-determinant term penalizes trivially large uncertainty estimates.
The learned uncertainty thus measures the reliability of cross-view pixel correspondences.

\subsection{Applications of Predicted Cross-View Uncertainty}
\label{sec:uncertainty-guided-optimization}

\methodblock{Dense-Point Filtering} For each pixel, we convert the diagonal covariance in Eq.~\eqref{eq:geometric-uncertainty-covariance} into a scalar uncertainty
\begin{equation}
    \sigma_i^{\mathrm{geo}}(\mathbf{u})
    =
    \sqrt{\operatorname{tr}\!\left(\boldsymbol{\Sigma}_i(\mathbf{u})\right)}
    =
    \sqrt{
        e^{\mathbf{S}_i^u(\mathbf{u})}
        +
        e^{\mathbf{S}_i^v(\mathbf{u})}
    }.
    \label{eq:scalar-geometric-uncertainty}
\end{equation}
To recover the dense point cloud, SURE-Map back-projects only pixels with valid predicted depths and sufficiently low uncertainty, \(\sigma_i^{\mathrm{geo}}(\mathbf{u})\). This filtering removes noisy points originating from pixels with high pose-depth-induced uncertainty.
Although the native streaming backbone also predicts depth confidence, this confidence does not reliably reflect the quality of the back-projected points, as demonstrated by the filtering ablation in \textcolor{blue}{Table~\ref{tab:filtering_ablation}} and \textcolor{blue}{Fig.~\ref{fig:filtering-ablation}}.

\methodblock{Uncertainty-Weighted Optimization} The same uncertainty weights point-to-plane residuals in local translation optimization to reduce pose errors from ambiguous observations.
We~keep predicted rotations fixed to prevent scale or correspondence residuals from inducing rotation errors that accumulate multiplicatively over long sequences.

For each pair of consecutive images \((\mathbf{I}_{i-1}, \mathbf{I}_i)\), let \(\mathbf{t}_{i \rightarrow i-1}\) denote the optimizable relative translation from \(\mathbf{I}_i\) to \(\mathbf{I}_{i-1}\). For notational simplicity, we write \(\mathbf{t}_{i \rightarrow i-1}\) as \(\mathbf{t}\) throughout this section when no ambiguity arises.

\begingroup
\displaywidowpenalty=10000
\predisplaypenalty=10000
Using the warping operation in Eq.~\eqref{eq:induced-flow}, a pixel \(\mathbf{u}\) with valid depth \(\mathbf{D}_i(\mathbf{u})\) can be warped from frame \(\mathbf{I}_i\) to the previous frame \(\mathbf{I}_{i-1}\). We denote the corresponding point in the previous frame by \(\mathbf{p}^{\mathrm{prev}}(\mathbf{u};\mathbf{t})\).
The residual \(r_i(\mathbf{u};\mathbf{t})\) denotes the point-to-plane distance between \(\mathbf{p}^{\mathrm{prev}}(\mathbf{u};\mathbf{t})\) and its corresponding local surface induced from the depth map \(\mathbf{D}_{i-1}\) :
\begin{equation}
    r_i(\mathbf{u};\mathbf{t})
    =
    \mathbf{n}^{\top}\!\left(
        \mathbf{p}^{\mathrm{prev}}(\mathbf{u};\mathbf{t})-\mathbf{q}
    \right),
    \label{eq:point-to-plane-residual}
\end{equation}
\endgroup
where \(\mathbf{q}\) is the surface point obtained by back-projecting the target pixel \(\mathbf{u}'=\pi\!\left(\mathbf{K}_{i-1}\mathbf{p}^{\mathrm{prev}}(\mathbf{u};\mathbf{t})\right)\) in \(\mathbf{I}_{i-1}\) using \(\mathbf{D}_{i-1}(\mathbf{u}')\) and \(\mathbf{K}_{i-1}\), and \(\mathbf{n}\) is its unit surface normal.

We further derive the uncertainty of the point-to-plane constraint from the predicted cross-view uncertainty, represented by the covariance \(\boldsymbol{\Sigma}_i(\mathbf{u})\) defined in Eq.~\eqref{eq:geometric-uncertainty-covariance}. Treating the predicted depth \(\mathbf{D}_{i-1}\) as locally fixed, we propagate \(\boldsymbol{\Sigma}_i(\mathbf{u})\) to the point-to-plane constraint via first-order Jacobian-based uncertainty propagation:
\begin{equation}
\begin{aligned}
    a_i^u(\mathbf{u})
    &=
    \mathbf{n}^{\top}\mathbf{R}_{i-1}^{\top}\mathbf{R}_i
    [\mathbf{D}_i(\mathbf{u})/f_{x,i},0,0]^\top,\\
    a_i^v(\mathbf{u})
    &=
    \mathbf{n}^{\top}\mathbf{R}_{i-1}^{\top}\mathbf{R}_i
    [0,\mathbf{D}_i(\mathbf{u})/f_{y,i},0]^\top,\\
    \mathbf{J}_i^r(\mathbf{u})
    &=
    \left[a_i^u(\mathbf{u}),a_i^v(\mathbf{u})\right],\\
    \sigma_{r,i}^2(\mathbf{u})
    &=
    \mathbf{J}_i^r(\mathbf{u})
    \boldsymbol{\Sigma}_i(\mathbf{u})
    \mathbf{J}_i^r(\mathbf{u})^\top
    + \sigma_0^2.
\end{aligned}
    \label{eq:residual-uncertainty-propagation}
\end{equation}
where \(a_i^u(\mathbf{u})\) and \(a_i^v(\mathbf{u})\) are partial derivatives of the point-to-plane residual with respect to horizontal and vertical current-frame pixel coordinates, and \(\mathbf{J}_i^r(\mathbf{u})=[a_i^u(\mathbf{u}),a_i^v(\mathbf{u})]\) is the corresponding residual-pixel Jacobian.
Here, \(\sigma_{r,i}^2(\mathbf{u})\) is the point-to-plane residual variance, \(f_{x,i}\) and \(f_{y,i}\) denote the camera focal lengths for \(\mathbf{I}_i\), and \(\sigma_0^2\) is a predefined residual-variance floor.

With the point-to-plane residual \(r_i(\mathbf{u};\mathbf{t})\) and its uncertainty \(\sigma_{r,i}^2(\mathbf{u})\), the objective function for optimizing the local translation is:
\begin{equation}
    \mathbf{t}^{\mathrm{GN}}_i
    =
    \arg\min_{\mathbf{t}}
    \sum_{\mathbf{u}\in\Omega_i^{g}}
    \frac{r_i(\mathbf{u};\mathbf{t})^2}{\sigma_{r,i}^2(\mathbf{u})},
    \label{eq:translation-objective}
\end{equation}
where \(\Omega_i^{g}\) denotes the set of pixels with valid projected geometry.
We solve the above nonlinear optimization problem using a small, fixed number of Gauss--Newton iterations. The initial value of the optimizable local translation $\mathbf{t}$ can be computed from the pose predictions of the streaming model.

\subsection{Multi-Timescale Self-Correction}
\label{sec:scale-calibration}

Consecutive-frame inference uses geometry-context-attention~\cite{chen2026geometric} for efficient causal updates, whereas full-attention enables all input frames to interact jointly and capture richer cross-frame geometric dependencies.

\methodblock{Keyframe-Window Inference} To preserve online efficiency, we maintain a fixed-size sliding keyframe window \(\mathcal{K}\).
Keyframes are selected and inserted at a fixed stride, with a new keyframe inserted when the magnitude of pose-depth-induced optical flow across the latest keyframe exceeds a predefined threshold.
We periodically run joint full-attention inference with the streaming model \(f_\theta\) over the current window:
\begin{equation}
    \{(\mathbf{D}_{c}^{k},\mathbf{T}_{c}^{k})\}_{c\in\mathcal{K}}
    =
    f_\theta(\{\mathbf{I}_c\}_{c\in\mathcal{K}}).
    \label{eq:keyframe-window-inference}
\end{equation}
where \(\mathbf{D}_c^k\) and \(\mathbf{T}_c^k\) denote the depth and pose for the frame $c$ predicted by the full-attention inference.
The depth map $\mathbf{D}_c^{k}$ obtained from keyframe-window inference (full-attention) will be aligned with its corresponding depth map $\mathbf{D}_c$ predicted by consecutive-frame inference (geometry-context-attention) via inverse-depth fitting:
\begin{equation}
    \eta
    =
    \arg\min_{\eta>0}
    \sum_{c\in\mathcal{K}}
    \sum_{\mathbf{u}\in\Omega_c}
    \left|
        \mathbf{D}_c(\mathbf{u})^{-1}
        -
        \eta\,\mathbf{D}_{c}^{k}(\mathbf{u})^{-1}
    \right|^2,
    \label{eq:segment-depth-alignment}
\end{equation}
where \(\Omega_c\) denotes the set of pixels with valid depth values in both depth maps.
\methodblock{Segment-Scale Estimation} The scale cue \(\phi\) is obtained by comparing keyframe-window relative translation lengths with the corresponding streaming relative translation lengths.
The keyframe-window relative translation lengths are divided by \(\eta\) to match the streaming-depth scale:
\begin{equation}
    \phi
    =
    \operatorname{median}_{(m,n)\in\mathcal{P}}
    \frac{
    \left\|
    \operatorname{trans}\!\left((\mathbf{T}_{m}^{k})^{-1}\mathbf{T}_{n}^{k}\right)
    \right\|/\eta
    }{
    \left\|
    \operatorname{trans}\!\left(\mathbf{T}_m^{-1}\mathbf{T}_n\right)
    \right\|
    },
    \label{eq:segment-scale-cue}
\end{equation}
where \(\mathcal{P}\) denotes valid pairs in the keyframe window and \(\operatorname{trans}(\cdot)\) extracts the translation vector.
\methodblock{Multi-Timescale Fusion} For each frame-to-frame relative edge in the recent trajectory segment covered by the current recalibration step, SURE-Map keeps the predicted rotation and updates only the translation.
We define the scale-recalibrated translation estimate as
\begin{equation}
    \mathbf{t}^{\mathrm{scale}}_i
    =
    \phi\,\operatorname{trans}(\mathbf{T}_{i-1}^{-1}\mathbf{T}_i).
    \label{eq:segment-scale-calibration}
\end{equation}
The final translation fuses the scale-recalibrated estimate with the local translation estimate from Eq.~\eqref{eq:translation-objective}:
\begin{equation}
    \mathbf{t}^{*}_i
    =
    \arg\min_{\mathbf{t}}
    \left[
    (1-\mu)\left\|\mathbf{t}-\mathbf{t}^{\mathrm{scale}}_i\right\|_2^2
    +
    \mu\left\|\mathbf{t}-\mathbf{t}^{\mathrm{GN}}_i\right\|_2^2
    \right],
    \label{eq:scale-gn-fusion}
\end{equation}
where \(\mu\) is a predefined hyperparameter that balances the scale-recalibrated translation estimate from keyframe-window inference and the local translation estimate from consecutive frames.

\raggedbottom
\section{Experiments}

\begin{table*}[!t]
    \centering
    \caption{ATE-RMSE (m) on KITTI~\cite{geiger2012we}. \bestscore{Red}/\secondscore{blue}: \bestscore{best}/\secondscore{second-best} within Streaming Fwd. LoGeR\textsuperscript{*}~\cite{zhang2026loger} denotes optimization-based LoGeR; CUT3R~\cite{wang2025continuous} and TTT3R~\cite{chen2026ttt3r} report both refresh settings. LC denotes loop closure.}
    \label{tab:kitti_ate}
    \vspace{0.2em}
    \setlength{\tabcolsep}{3.0pt}
    \renewcommand{\arraystretch}{0.80}
    \resizebox{\textwidth}{!}{
    \begin{tabular}{@{}ll|rrrrrrrrrrr|r@{}}
        \toprule
        \multicolumn{2}{l|}{\textbf{Methods}} &
        \multicolumn{11}{c|}{\textbf{KITTI ATE} \(\downarrow\)} &
        \textbf{Avg.} \\
        \cmidrule(lr){3-13}
        & &
        \textbf{00} & \textbf{01} & \textbf{02} & \textbf{03} & \textbf{04} &
        \textbf{05} & \textbf{06} & \textbf{07} & \textbf{08} & \textbf{09} & \textbf{10} & \\
        & &
        \textcolor{blue!70!black}{4542 fr.} &
        \textcolor{blue!70!black}{1101 fr.} &
        \textcolor{blue!70!black}{4661 fr.} &
        \textcolor{blue!70!black}{801 fr.} &
        \textcolor{blue!70!black}{271 fr.} &
        \textcolor{blue!70!black}{2761 fr.} &
        \textcolor{blue!70!black}{1101 fr.} &
        \textcolor{blue!70!black}{1101 fr.} &
        \textcolor{blue!70!black}{4071 fr.} &
        \textcolor{blue!70!black}{1591 fr.} &
        \textcolor{blue!70!black}{1201 fr.} & \\
        & &
        \textcolor{red!75!black}{3.7 km} &
        \textcolor{red!75!black}{2.5 km} &
        \textcolor{red!75!black}{5.1 km} &
        \textcolor{red!75!black}{0.6 km} &
        \textcolor{red!75!black}{0.4 km} &
        \textcolor{red!75!black}{2.2 km} &
        \textcolor{red!75!black}{1.2 km} &
        \textcolor{red!75!black}{0.7 km} &
        \textcolor{red!75!black}{3.2 km} &
        \textcolor{red!75!black}{1.7 km} &
        \textcolor{red!75!black}{0.9 km} & \\
        \midrule
        \multirow{6}{*}{\rotatebox{90}{Opt.-centric}} &
        MASt3R-SLAM~\cite{murai2025mast3r} & -- & 530.37 & -- & 18.87 & 88.98 & 159.43 & 92.00 & -- & 263.75 & -- & 153.07 & 186.64 \\
        & VGGT-SLAM 2.0~\cite{maggio2026vggtslam2} & -- & 163.65 & -- & 50.04 & 19.38 & 159.58 & 46.35 & 57.80 & -- & 167.96 & 76.99 & 92.72 \\
        & COLMAP~\cite{schonberger2016structure} & 139.12 & 3.83 & 71.99 & 1.46 & 112.77 & 20.37 & 10.95 & 7.80 & 21.72 & 21.19 & 4.52 & 37.79 \\
        & MASt3R-SfM~\cite{duisterhof2025mast3r} & -- & 463.52 & -- & 15.80 & 41.44 & 150.39 & 136.14 & 71.69 & -- & 176.36 & 69.50 & 140.60 \\
        & DPVO~\cite{teed2023deep} & 113.11 & 16.60 & 113.01 & 2.46 & 0.98 & 59.34 & 55.91 & 19.30 & 110.63 & 74.55 & 13.71 & 52.69 \\
        & DROID-SLAM~\cite{teed2021droid} & -- & 82.81 & -- & 3.20 & 1.47 & 73.50 & 61.10 & 18.41 & 104.22 & 89.49 & 22.19 & 50.71 \\
        \midrule
        \multirow{5}{*}{\rotatebox{90}{Offline Fwd.}} &
        VGGT-Long~\cite{deng2025vggt} & 8.64 & 61.21 & 52.72 & 8.78 & 4.20 & 9.88 & 4.67 & 2.66 & 72.98 & 31.84 & 27.71 & 25.94 \\
        & FastVGGT~\cite{shen2025fastvggt} & -- & 639.39 & -- & 21.53 & 9.51 & -- & 40.56 & 51.35 & -- & 201.54 & 196.22 & 165.73 \\
        & LoGeR~\cite{zhang2026loger} & 62.34 & 41.64 & 39.64 & 4.89 & 1.82 & 41.27 & 13.99 & 16.24 & 26.46 & 22.71 & 8.84 & 25.44 \\
        & LoGeR\textsuperscript{*}~\cite{zhang2026loger} & 30.47 & 47.91 & 36.32 & 5.38 & 1.95 & 26.34 & 6.60 & 5.55 & 24.41 & 10.12 & 10.11 & 18.65 \\
        & Scal3R~\cite{xie2026scal3r} & 4.30 & 45.29 & 42.06 & 3.36 & 1.74 & 3.30 & 2.49 & 2.03 & 36.69 & 12.32 & 6.46 & 14.55 \\
        \midrule
        \multirow{13}{*}{\rotatebox{90}{Streaming Fwd.}} &
        CUT3R~\cite{wang2025continuous} w/o refresh & 185.89 & 651.52 & 296.98 & 148.06 & 22.17 & 155.61 & 132.54 & 77.03 & 238.39 & 205.94 & 193.39 & 209.78 \\
        & CUT3R~\cite{wang2025continuous} w/ refresh & 190.38 & 90.59 & 264.39 & 20.40 & 7.31 & 92.25 & 67.54 & 22.48 & 145.08 & 67.42 & 40.00 & 91.62 \\
        & TTT3R~\cite{chen2026ttt3r} w/o refresh & 190.93 & 546.84 & 218.77 & 105.28 & 11.62 & 153.12 & 132.94 & 70.95 & 180.57 & 211.01 & 133.00 & 177.73 \\
        & TTT3R~\cite{chen2026ttt3r} w/ refresh & 119.94 & 99.59 & 238.07 & 16.83 & 3.98 & 36.38 & 47.20 & 11.62 & 107.33 & 86.96 & 33.58 & 72.86 \\
        & Stream3R~\cite{lan2026stream3r} & 190.98 & 681.95 & 301.40 & 158.25 & 102.73 & 159.85 & 135.03 & 90.37 & 261.15 & 216.31 & 207.49 & 227.77 \\
        & StreamVGGT~\cite{zhuo2026streaming} & 191.93 & 653.06 & 303.35 & 157.50 & 108.24 & 160.46 & 133.71 & 89.00 & 263.95 & 216.69 & 209.80 & 226.15 \\
        & InfiniteVGGT~\cite{yuan2026infinitevggt} & 186.46 & 623.62 & 289.16 & 166.74 & 68.00 & 143.84 & 117.57 & 85.33 & 221.56 & 215.41 & 156.92 & 206.78 \\
        & LongStream~\cite{cheng2026longstream} & 92.55 & 46.01 & 134.70 & 3.81 & 1.95 & 84.69 & 23.12 & 14.93 & 62.07 & 85.61 & 21.48 & 51.90 \\
        & LingBot-Map~\cite{chen2026geometric} & 23.09 & 88.05 & 70.57 & \secondscore{2.80} & \secondscore{0.95} & 18.48 & 4.90 & \secondscore{5.03} & \bestscore{13.83} & 22.17 & 14.20 & 24.00 \\
        & HorizonStream~\cite{cheng2026horizonstream} & 29.20 & \bestscore{23.90} & 96.98 & 5.56 & \bestscore{0.71} & \secondscore{13.97} & 5.41 & 6.19 & 22.32 & 29.58 & \secondscore{12.29} & 22.38 \\
        & HorizonStream~\cite{cheng2026horizonstream} w/LC & \bestscore{15.37} & \bestscore{23.90} & 70.89 & 5.56 & \bestscore{0.71} & \bestscore{7.93} & 7.72 & 6.19 & 22.32 & 27.18 & \secondscore{12.29} & 18.19 \\
        & SURE-Map & 20.81 & \secondscore{39.54} & \secondscore{53.47} & \bestscore{2.28} & 1.33 & 18.44 & \secondscore{4.00} & \bestscore{3.99} & \secondscore{14.38} & \secondscore{21.15} & \bestscore{10.20} & \secondscore{17.24} \\
        & SURE-Map w/LC & \secondscore{20.59} & \secondscore{39.54} & \bestscore{35.85} & \bestscore{2.28} & 1.33 & 17.48 & \bestscore{3.36} & \bestscore{3.99} & \secondscore{14.38} & \bestscore{17.87} & \bestscore{10.20} & \bestscore{15.17} \\
        \bottomrule
    \end{tabular}}
    \vspace{0.2em}
\end{table*}

\begin{table}[!t]
    \centering
    \captionsetup{font=small}
    \caption{Average ATE-RMSE (m) on all sequences of the long-horizon benchmarks.}
    \label{tab:pose_summary}
    \vspace{-0.05em}
    \setlength{\tabcolsep}{3.0pt}
    \renewcommand{\arraystretch}{0.70}
    \resizebox{0.90\columnwidth}{!}{
    \begin{tabular}{@{}lccc@{}}
        \toprule
        \textbf{Method} & \textbf{KITTI~\cite{geiger2012we}} & \textbf{Oxford~\cite{tao2026oxford}} & \textbf{VBR~\cite{brizi2024vbr}} \\
        \midrule
        CUT3R~\cite{wang2025continuous} & 91.62 & 32.47 & 66.25 \\
        TTT3R~\cite{chen2026ttt3r} & 72.86 & 25.05 & 64.99 \\
        InfiniteVGGT~\cite{yuan2026infinitevggt} & 206.78 & 31.75 & 91.60 \\
        LongStream~\cite{cheng2026longstream} & 51.90 & 15.92 & 77.93 \\
        LingBot-Map~\cite{chen2026geometric} & 24.00 & 5.11 & 31.37 \\
        HorizonStream~\cite{cheng2026horizonstream} & 22.38 & 8.87 & 29.68 \\
        HorizonStream~\cite{cheng2026horizonstream} w/LC & 18.19 & 8.12 & 22.87 \\
        SURE-Map & 17.24 & 4.74 & 28.58 \\
        SURE-Map w/LC & \textbf{15.17} & \textbf{4.63} & \textbf{22.12} \\
        \bottomrule
\end{tabular}}
    \vspace{0.2em}
\end{table}

\subsection{Datasets and Implementation Details}
\label{sec:datasets}

We train only the cross-view geometric uncertainty head on TartanAir~\cite{wang2020tartanair}, using ground-truth optical flow and validity masks for residual-based supervision (see Eq.~\eqref{eq:flow-uncertainty-loss}).
We use a frozen LingBot-Map~\cite{chen2026geometric} backbone and train the uncertainty head for \(20\mathrm{k}\) iterations with the NLL loss in Sec.~\ref{sec:uncertainty-learning}.
Training uses \(8\)--\(24\)-frame clips at \(518\times392\) and AdamW with a learning rate of \(10^{-4}\).

For evaluation, we assess long-horizon trajectory accuracy on KITTI~\cite{geiger2012we}, VBR~\cite{brizi2024vbr}, and Oxford Spires~\cite{tao2026oxford}, and evaluate dense reconstruction with geometric-uncertainty filtering on Neural RGB-D~\cite{azinovic2022neural} and 7-Scenes~\cite{shotton2013scene}.

\subsection{Baselines}

For long-horizon trajectory evaluation, we group baselines by their dominant estimation paradigm: optimization-centric SfM/SLAM methods~\cite{murai2025mast3r,maggio2026vggtslam2,schonberger2016structure,duisterhof2025mast3r,teed2021droid,teed2023deep}, offline feed-forward methods~\cite{deng2025vggt,zhang2026loger,xie2026scal3r,shen2025fastvggt}, and streaming feed-forward methods~\cite{chen2026ttt3r,cheng2026longstream,wang2025continuous,lan2026stream3r,chen2026geometric,cheng2026horizonstream,zhuo2026streaming,yuan2026infinitevggt}, as summarized in Tables~\ref{tab:kitti_ate} and~\ref{tab:pose_summary}.
Optimization-centric methods may operate online but rely primarily on iterative pose estimation.
For loop closure (LC), we retrieve revisited frame pairs with cached early-layer DINOv2~\cite{oquab2023dinov2} features and use the resulting geometric corrections as pose graph optimization (PGO) constraints.
For point-cloud filtering evaluation, we compare against streaming feed-forward reconstruction methods~\cite{chen2026ttt3r,wang2025continuous,lan2026stream3r,chen2026geometric,cheng2026horizonstream,zhuo2026streaming,yuan2026infinitevggt,li2026wint3r} on Neural RGB-D and 7-Scenes.

\subsection{Evaluation Metrics}

For trajectory evaluation, we report ATE-RMSE~\cite{sturm2012benchmark} in meters after standard \(\mathrm{Sim}(3)\) trajectory alignment.
For dense reconstruction, we report Accuracy, Completeness, Chamfer Distance (CD), and F1 score after point-cloud alignment~\cite{zuo2023incremental}; F1 uses a \(0.05\,\mathrm{m}\) distance threshold.

\begin{table}[!t]
    \centering
    \captionsetup{position=top,skip=3pt}
    \caption{Dense reconstruction results for streaming feed-forward methods. Acc./Comp.: m; F1: \%.}
    \label{tab:dense_reconstruction}
    \setlength{\tabcolsep}{3.2pt}
    \renewcommand{\arraystretch}{0.90}
    \resizebox{0.88\columnwidth}{!}{
    \begin{tabular}{@{}lcccccc@{}}
        \toprule
        \multirow{2}{*}{\textbf{Method}} & \multicolumn{3}{c}{\textbf{NRGBD}~\cite{azinovic2022neural}} & \multicolumn{3}{c}{\textbf{7-Scenes}~\cite{shotton2013scene}} \\
        \cmidrule(lr){2-4}\cmidrule(lr){5-7}
        & \textbf{Acc.} \(\downarrow\) & \textbf{Comp.} \(\downarrow\) & \textbf{F1} \(\uparrow\) & \textbf{Acc.} \(\downarrow\) & \textbf{Comp.} \(\downarrow\) & \textbf{F1} \(\uparrow\) \\
        \midrule
        StreamVGGT~\cite{zhuo2026streaming} & 0.13 & 0.05 & 45.08 & 0.04 & 0.11 & 69.44 \\
        InfiniteVGGT~\cite{yuan2026infinitevggt} & 0.13 & 0.05 & 42.27 & 0.04 & 0.11 & 68.53 \\
        CUT3R~\cite{wang2025continuous} & 0.25 & 0.15 & 32.22 & 0.07 & 0.10 & 58.98 \\
        TTT3R~\cite{chen2026ttt3r} & 0.16 & 0.06 & 53.55 & 0.03 & 0.08 & 77.25 \\
        Wint3R~\cite{li2026wint3r} & 0.09 & 0.04 & 56.96 & 0.03 & 0.07 & 78.81 \\
        Stream3R~\cite{lan2026stream3r} & 0.21 & 0.07 & 54.07 & \textbf{0.02} & 0.09 & 78.79 \\
        HorizonStream~\cite{cheng2026horizonstream} & 0.14 & 0.07 & 49.13 & 0.79 & 0.30 & 51.09 \\
        LingBot-Map~\cite{chen2026geometric} & 0.074 & \textbf{0.030} & 65.10 & 0.035 & \textbf{0.043} & 81.77 \\
        SURE-Map & \textbf{0.067} & 0.035 & \textbf{66.20} & 0.033 & 0.045 & \textbf{81.93} \\
        \bottomrule
    \end{tabular}}
    \vspace{-0.6em}
\end{table}

\begin{lrbox}{\poseablationbox}
    \begin{minipage}{\columnwidth}
        \centering
        \captionsetup{font=footnotesize}
        \captionof{table}{Pose-estimation ablation with ATE-RMSE (m).}
        \label{tab:pose_ablation}
        \vspace{0.05em}
        \footnotesize
        \setlength{\tabcolsep}{3.5pt}
        \renewcommand{\arraystretch}{0.72}
        \resizebox{0.92\columnwidth}{!}{
        \begin{tabular}{@{}lccc@{}}
            \toprule
            \textbf{Variant} & \textbf{Oxford~\cite{tao2026oxford}} & \textbf{KITTI~\cite{geiger2012we}} & \textbf{VBR~\cite{brizi2024vbr}} \\
            \midrule
            SURE-Map w/LC & \textbf{4.63} & \textbf{15.17} & \textbf{22.12} \\
            $-$ LC & 4.74 & 17.24 & 28.58 \\
            $-$ uncertainty-weighted opt. & 4.91 & 17.57 & 29.96 \\
            $-$ scale recalibration & 5.11 & 24.00 & 31.37 \\
            \bottomrule
        \end{tabular}}
    \end{minipage}
\end{lrbox}

\begin{figure}[!t]
    \centering
    \includegraphics[width=\columnwidth,trim=25bp 19bp 12bp 26bp,clip]{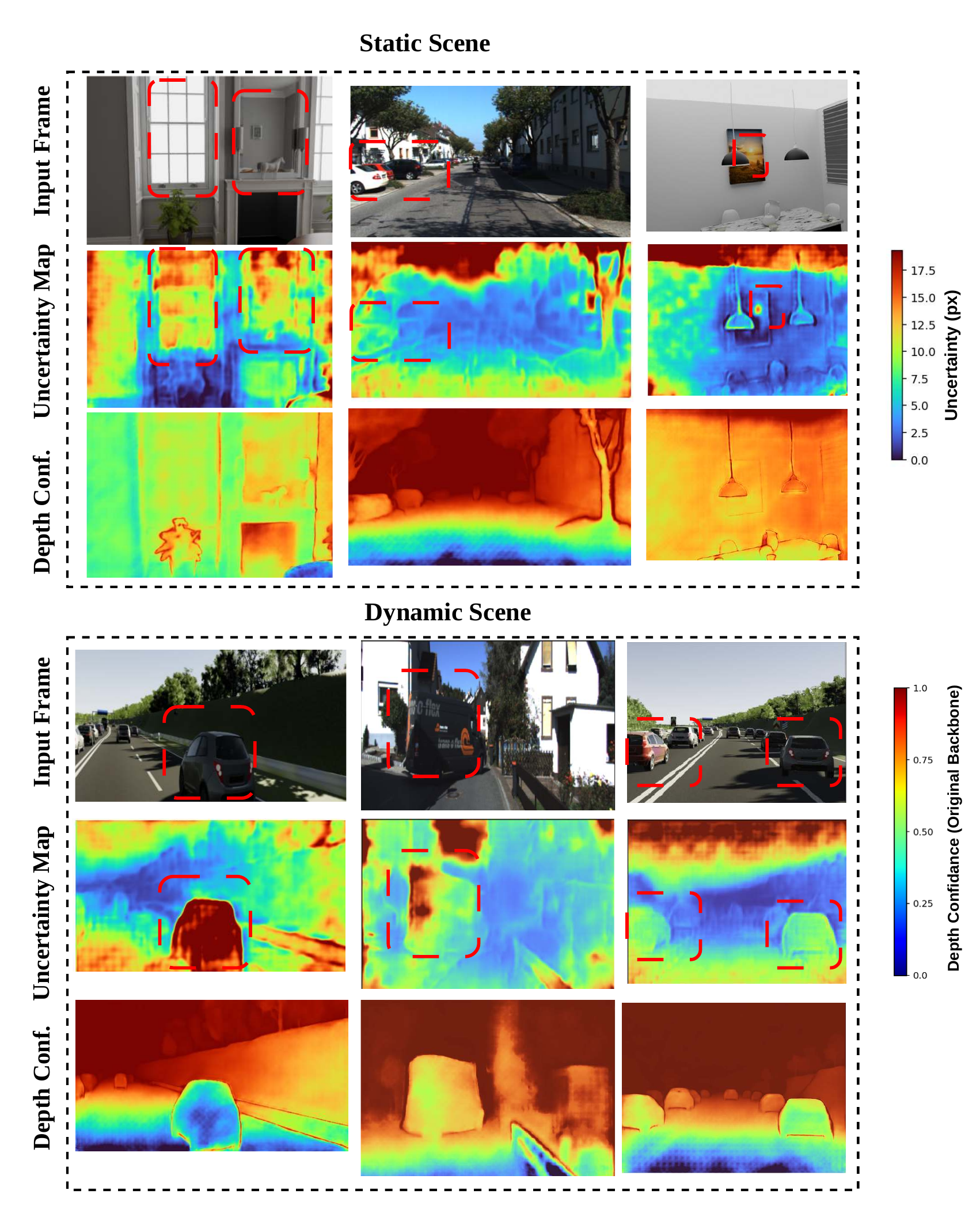}
    \vspace{-0.5em}
    \caption{
        Comparison of our learned cross-view uncertainty and the depth confidence from the streaming model~\cite{chen2026geometric}.
        Warmer colors indicate higher uncertainty.
        Our geometric uncertainty map highlights dynamic or ambiguous regions where pose-depth-induced optical flow disagrees with consistent pixel correspondences while preserving static vehicles; depth confidence mainly reflects range-dependent reliability.
    }
    \label{fig:uncertainty-visualization}
    \vspace{0.8em}
    \begin{minipage}{\columnwidth}
        \centering
        \captionsetup{font=footnotesize,position=top,skip=6pt,belowskip=-4pt}
        \captionof{table}{Point-cloud filtering ablation. CD/Comp.: m; F1: \%.}
        \label{tab:filtering_ablation}
        \vspace{0.05em}
        \footnotesize
        \setlength{\tabcolsep}{2.0pt}
        \renewcommand{\arraystretch}{0.72}
        \resizebox{\columnwidth}{!}{
        \begin{tabular}{@{}llccc@{}}
            \toprule
            \textbf{Dataset} & \textbf{Variant} & \textbf{CD} \(\downarrow\) & \textbf{Comp.} \(\downarrow\) & \textbf{F1} \(\uparrow\) \\
            \midrule
            \multirow{3}{*}{NRGBD~\cite{azinovic2022neural}} & SURE-Map & \textbf{0.051} & 0.035 & \textbf{66.20} \\
            & SURE-Map w/o uncertainty filtering & 0.052 & \textbf{0.030} & 65.10 \\
            & SURE-Map w/ depth conf. filtering & 0.084 & 0.113 & 65.00 \\
            \midrule
            \multirow{3}{*}{7-Scenes~\cite{shotton2013scene}} & SURE-Map & \textbf{0.039} & 0.045 & \textbf{81.93} \\
            & SURE-Map w/o uncertainty filtering & \textbf{0.039} & \textbf{0.043} & 81.77 \\
            & SURE-Map w/ depth conf. filtering & 0.051 & 0.079 & 80.90 \\
            \bottomrule
        \end{tabular}}
    \end{minipage}
\end{figure}

\begin{figure}[t]
    \centering
    \includegraphics[width=\columnwidth,height=1.035\columnwidth,keepaspectratio,trim=36bp 20bp 19bp 37bp,clip]{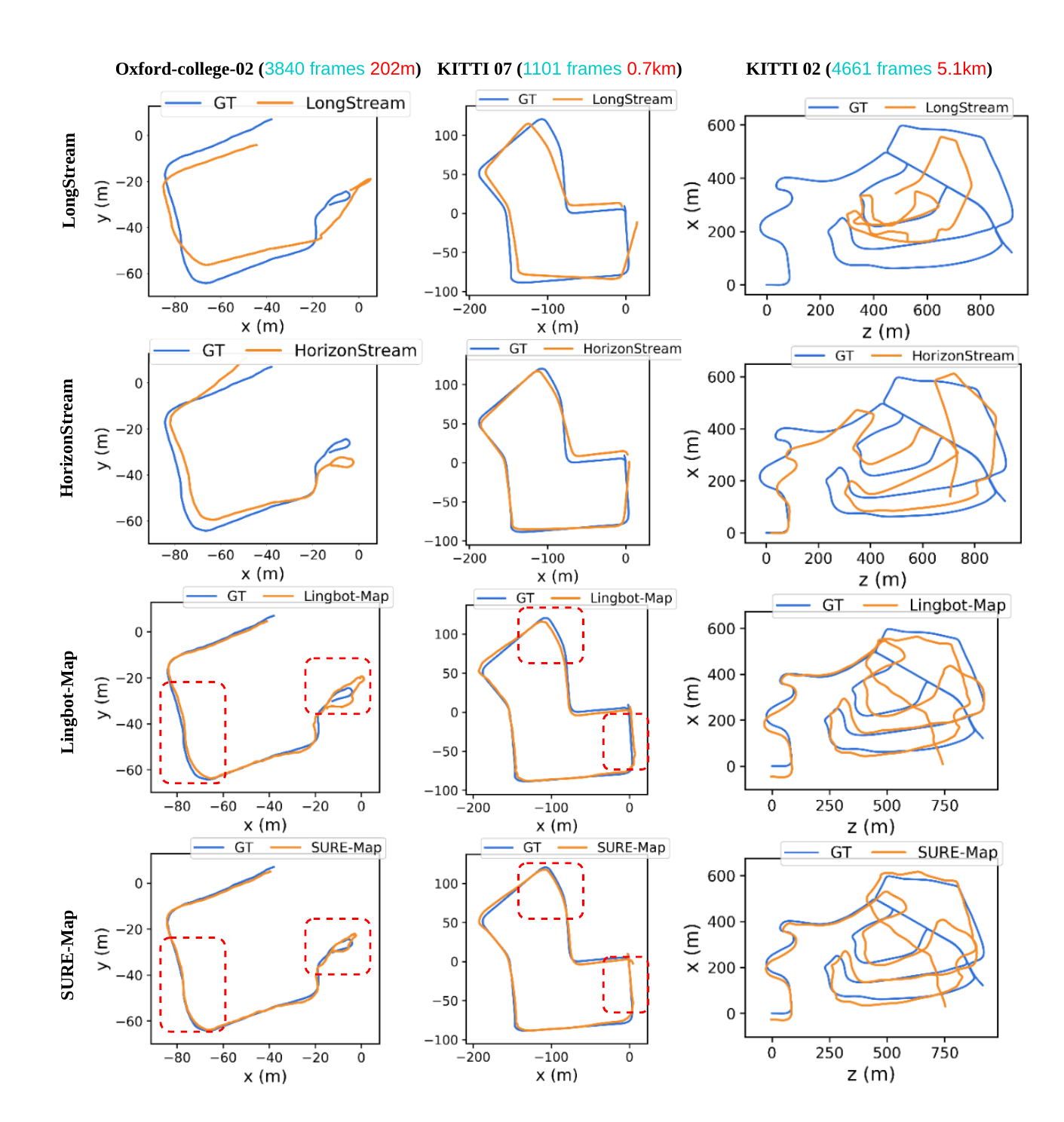}
    \vspace{-0.8em}
    \caption{
        Qualitative trajectory comparison on long-horizon sequences. SURE-Map better preserves global trajectory shape compared to LingBot-Map~\cite{chen2026geometric}, HorizonStream~\cite{cheng2026horizonstream}, and LongStream~\cite{cheng2026longstream}. \textcolor{blue}{\textbf{Ground truth}} is shown in blue, and \textcolor{orange}{\textbf{predictions}} are shown in orange.
    }
    \label{fig:trajectory-qualitative}
    \vspace{-0.6em}
\end{figure}

\subsection{Experimental Results}

\figref{fig:uncertainty-visualization} compares our learned cross-view geometric uncertainty with the depth confidence predicted from the streaming 3D backbone~\cite{chen2026geometric}.
Our cross-view uncertainty highlights mismatches between pose-depth-induced optical flow and the cross-frame pixel correspondences, often in dynamic or ambiguous regions.
Depth confidence instead primarily reflects the range-dependent reliability of individual-view geometry rather than cross-view correspondence quality.

\methodblock{Camera Pose Estimation} \figref{fig:trajectory-qualitative} compares representative long-horizon trajectories without PGO post-processing, showing that SURE-Map reduces segment-wise drift through online correction alone.
Table~\ref{tab:kitti_ate} reports per-sequence KITTI~\cite{geiger2012we} results, while Table~\ref{tab:pose_summary} summarizes average ATE-RMSE on KITTI, Oxford Spires~\cite{tao2026oxford}, and VBR~\cite{brizi2024vbr}.
SURE-Map performs strongly across all benchmarks.
Optional PGO incorporates LC constraints between revisited frames, further improving global consistency.

\begingroup
\clubpenalty=0
\widowpenalty=0
\interlinepenalty=0
\methodblock{3D Reconstruction with Geometric-Uncertainty Filtering} Neural RGB-D~\cite{azinovic2022neural} features cluttered indoor scenes, and 7-Scenes~\cite{shotton2013scene} provides clean indoor environments.
Table~\ref{tab:dense_reconstruction} shows that geometric-uncertainty filtering improves accuracy and F-score while largely preserving completeness.
\par
\endgroup

\methodblock{Runtime Analysis} Under the same streaming setting, SURE-Map adds \(15\)--\(30\,\mathrm{ms}\) per frame over LingBot-Map~\cite{chen2026geometric}.
On KITTI sequence 04~\cite{geiger2012we}, throughput decreases from \(11.68\) to \(9.17\) FPS. Despite the slight FPS reduction, SURE-Map still preserves streaming efficiency.

\begin{figure}[!t]
    \centering
    \captionsetup{font=normalsize}
    \includegraphics[width=\columnwidth,trim=12pt 18pt 12pt 8pt,clip]{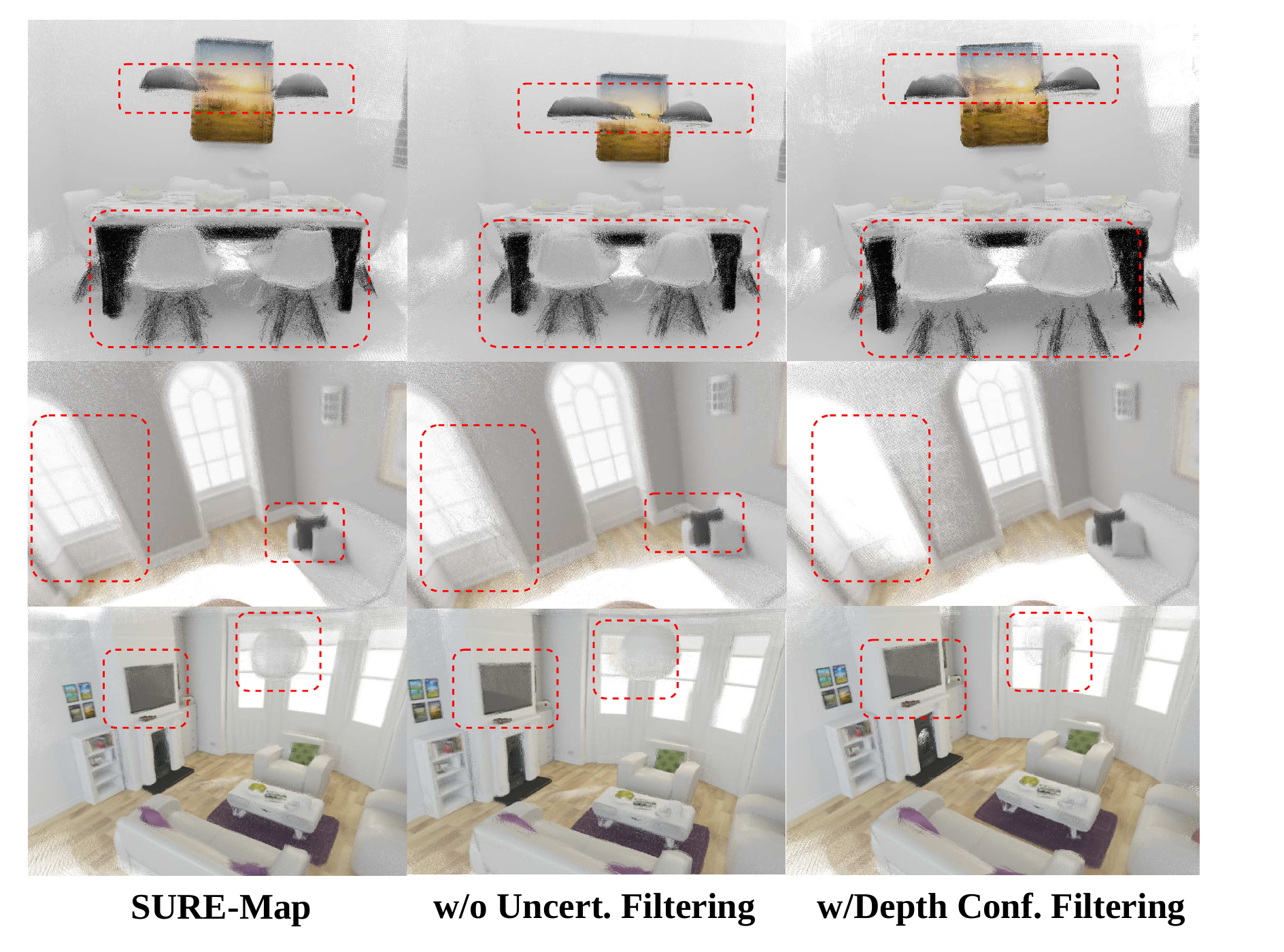}
    \vspace{0.2em}
    \caption{
        Qualitative point-cloud filtering ablation. Learned geometric uncertainty rejects unreliable points while preserving valid scene structures.
    }
    \label{fig:filtering-ablation}
    \vspace{0.8em}
    \usebox{\poseablationbox}
\end{figure}

\enlargethispage{\baselineskip}
\Needspace{3\baselineskip}
\methodblock{Ablation Studies} Table~\ref{tab:pose_ablation} reports progressive ablations of scale recalibration (Sec.~\ref{sec:scale-calibration}), uncertainty-weighted optimization (Sec.~\ref{sec:uncertainty-guided-optimization}), and loop closure (LC).
Scale recalibration reduces accumulated drift, while uncertainty-weighted optimization provides complementary local pose corrections.
At a fixed point-removal ratio, Table~\ref{tab:filtering_ablation} and Fig.~\ref{fig:filtering-ablation} show that uncertainty filtering improves geometric quality while preserving valid structures, whereas depth-confidence filtering discards more valid points and degrades reconstruction.

\FloatBarrier

\section{Conclusion}

\begingroup
\clubpenalty=0
\widowpenalty=0
\interlinepenalty=0
We presented SURE-Map, a self-correcting framework for streaming geometric foundation models.
It uses cross-view geometric uncertainty to identify unreliable pixel correspondences for indoor point-cloud filtering, while multi-timescale self-correction couples fast consecutive-frame inference with sparse keyframe-window inference to correct long-horizon trajectory errors.
Experiments demonstrate state-of-the-art long-horizon trajectory accuracy and improved dense geometry while preserving streaming efficiency.
\par
\endgroup

\FloatBarrier

\bibliographystyle{ieeetr}
\bibliography{ref}

\end{document}